# A New Robot Arm Calibration Method Based on Cubic Interpolated Beetle Antennae Search Approach

Zhibin Li, Shuai Li, *Senior Member*, *IEEE*, and Xin Luo, *Senior Member*, *IEEE*

*Abstract*—Industrial robot arms are extensively important for intelligent manufacturing. An industrial robot arm commonly enjoys its high repetitive positioning accuracy while suffering from its low absolute positioning accuracy, which greatly restricts its application in high-precision manufacture, like automobile manufacture. Aiming at addressing this hot issue, this work proposes a novel robot arm calibration method based on cubic interpolated beetle antennae search (CIBAS). This study has three ideas: a) developing a novel CIBAS algorithm, which can effectively addresses the issue of local optimum in a Beetle Antennae Search algorithm; b) utilizing a particle filter to reduce the influence of non-Gaussian noises; and c) proposing a new calibration method incorporating CIBAS algorithm and particle filter to searching the optimal kinematic parameters. Experimental results on an ABB IRB120 industrial robot arm demonstrate that the proposed method achieves much higher calibration accuracy than several state-of-the-art calibration methods.

*Index Terms*—Evolutionary Computation, Industrial Robot Arm, Robot Calibration, Absolute Positioning Accuracy, Particle Filter, Cubic Interpolated Beetle Antennae Search, Optimization.

## I. Introduction

WITH THE RAPID PROGRESS of industrial internet, various advanced manufacture-related issues emerge, such as automobile assembly and aircraft manufacturing [1-5]. Recently, owing to its high flexibility and universality, industrial robot arms are becoming a critical category of production equipment in intelligent manufacture environment [6-10].

Due to the inevitable impact factors such as manufacturing tolerance [15-20], a robot arm is apt to suffer from position error. Hence, how to improve the absolute positioning accuracy of a robot arm becomes a thorny issue. To date, researchers have proposed numerous calibration methods for improving the absolute positioning accuracy of a robot arm. Wang *et al.* [2] incorporate the principle of particle swarm optimization into a calibration method for a surgical robot's geometric parameters. Wu *et al.* [4] develop a framework for calibrating the multi-constraint parallel continuum robot with tip position data, which successfully reduces its position error by using Levenberg-Marquardt (LM) algorithm and interior point method. Wen *et al.* [6] utilize an improved crow search algorithm to identify the kinematic parameters of a Staubli-TX60 robot arm, thereby greatly improving its pose accuracy. The algorithms mentioned above facilitate the robot arm calibration to improve its absolute positioning accuracy [21-25]. However, they suffer from slow convergence and low calibration accuracy.

A Beetle Antennae Search (BAS) algorithm is a recent evolutionary computing algorithm that imitates the foraging behavior of a beetle to achieve the optimal solution for a complex optimization problem [11-15]. Compared with other evolutionary computing algorithms like particle swarm optimization (PSO) or genetic algorithm (GA), a BAS algorithm maintains an individual particle only to find the optimal solution without recognizing a detailed objective function and its gradient information. Hence, it has the advantages of low computation, fast convergence and ease of implementation [31-35].

Motivated by the above virtues of a BAS algorithm, this paper aims to establish efficient and accurate robot arm calibration based on its principle. However, a BAS algorithm frequently encounters local optimum and instability, which can greatly harm the calibration accuracy. To further address these issues, this paper proposes a novel Cubic Interpolated Beetle Antennae Search (CIBAS)-based robot arm calibration method.

This paper achieves the following significant contributions:
a) A novel CIBAS algorithm, which effectively addresses the issue of local optimum and unstable searching process in a conventional BAS algorithm. Its convergence is rigorously proved in theory;
b) A particle filter (PF) is employed to suppress the non-Gaussian noises;
c) A novel robot arm calibration method based on PF and CIBAS, which achieves high accuracy and efficiency.

The remainder of this paper is organized into four sections. Section II describes the preliminaries. Section III presents the PF and CIBAS algorithm for robot arm calibration. Section IV provides the experimental results and comparisons. Finally, the conclusions and future work are drawn in Section V.

✧ Z. Li and X. Luo are with the School of Computer Science and Technology, Chongqing University of Posts and Telecommunications, Chongqing 400065, China (e-mail: LiZhibin111@outlook.com, luoxin21@gmail.com).
✧ S. Li and X, Luo are with Zienkiewicz Centre for Computational Engineering and Department of Mechanical Engineering, College of Engineering, Swansea University Bay Campus, Swansea SA1 8EN, UK (email: shuai.li@swansea.ac.uk, luoxin21@gmail.com).

## II. PRELIMINARIES

### A. Kinematic and Error Model

Generally, the most classical kinematic model of industrial robot arm is DH model, which can accurately express the pose of the robot arm end-effector [36-40]. According to the DH rule, the transformation matrix of robotic adjacent joints is as follows.

$$^{i-1}T_i = R(z,\theta_i) \times T(z,d_i) \times T(x,a_i) \times R(x,\alpha_i)$$

$$= \begin{bmatrix} c\theta_i & -s\theta_i c\alpha_i & s\theta_i s\alpha_i & a_i c\theta_i \\ s\theta_i & c\theta_i c\alpha_i & -c\theta_i s\alpha_i & a_i s\theta_i \\ 0 & s\alpha_i & c\alpha_i & d_i \\ 0 & 0 & 0 & 1 \end{bmatrix}, \quad (1)$$

where $c\theta = cos\theta$, $s\theta = sin\theta$, $^{i-1}T_i$ is the link transformation matrix. $a_i$, $d_i$, $\theta_i$ and $\alpha_i$ are link length, link offset, joint angle, link twist angle, respectively.

Then, the kinematic model of the robot arm can be expressed as the following equation.

$$T_6 = {}^0T_1 {}^1T_2 {}^2T_3 {}^3T_4 {}^4T_5 {}^5T_6, \quad (2)$$

Due to the error of each kinematic parameter causes the link transformation matrix error of robotic adjacent joints, which can be shown in the following equation.

$$d\,{}^{i-1}T_i = \frac{\partial\,{}^{i-1}T_i}{\partial a_i}\Delta a_i + \frac{\partial\,{}^{i-1}T_i}{\partial d_i}\Delta d_i + \frac{\partial\,{}^{i-1}T_i}{\partial \theta_i}\Delta \theta_i + \frac{\partial\,{}^{i-1}T_i}{\partial \alpha_i}\Delta \alpha_i, \quad (3)$$

With (3), the relationship between the position deviation and kinematic parameter deviation can be described as

$$\Delta P = \begin{bmatrix} d_x \\ d_y \\ d_z \end{bmatrix} = \begin{bmatrix} J_1 & J_2 & J_3 & J_4 \end{bmatrix} \begin{bmatrix} \Delta \alpha \\ \Delta a \\ \Delta d \\ \Delta \theta \end{bmatrix} = J \cdot w, \quad (4)$$

where $\Delta P$ represents the position errors vector of the robot arm end-effector in $x$, $y$, and $z$, $J$ is Jacobian matrix of DH model, $\Delta \alpha$, $\Delta a$, $\Delta d$, and $\Delta \theta$ are the kinematic parameter deviations of the robot arm.

Based on the robot arm kinematic model and error model, we can approximately calculate the robot arm kinematic parameter deviations through the measuring cable length $L_i$ and the nominal cable length $L'_i$, then its fitness function is given as

$$f(w) = \min\left[\frac{1}{n}\sum_{i=1}^{n}(L_i - L'_i)^2\right], \quad (5)$$

where $n$ is the number of samples. $L'_i$ can be obtained by the following equation.

$$L'_i = \sqrt{(P'_i - P'_0)^2}. \quad (6)$$

$P'_i$ and $P'_0$ represent the nominal position of robot arm end-effector and the calculated position of fixed point, respectively.

## III. ACCURATE IDENTIFICATION OF KINEMATIC PARAMETERS BASED ON PF AND CIBAS ALGORITHM

### A. CIBAS Algorithm

The BAS algorithm is a novel intelligent optimization algorithm proposed by Jiang et al. [14], which is widely employed in various fields. To simply describe the principle of BAS algorithm, we give the following random direction of beetle searching.

$$\vec{b} = \frac{rands(k,1)}{\|rands(k,1)\|_2}, \quad (7)$$

where $\vec{b}$ and rands($\cdot$) denote a random direction of beetle searching and a random function, respectively. Moreover, the positions of the left and right tentacles of the beetle are presented as follows.

$$\begin{cases} w_l = w_t + m_0\vec{b}, \\ w_r = w_t - m_0\vec{b}. \end{cases} \quad (8)$$

where $w_t$ is the position of the beetle, $m_0$ represents the distance between the two left and right tentacles of the beetle. Considering the searching behavior of the beetle, we can obtain the following update formula of its position.

$$w_{t+1} = w_t + \delta_t \vec{b}\, sign(f(w_r) - f(w_l)), \quad (9)$$

With (9), $\delta$, $f$ and sign($\cdot$) are the step size of searching, the fitness function and the sign function, respectively.

According to the step update frequency of the BAS algorithm, the corresponding step $\delta_t$ updating equation is computed by

$$\delta_{t+1} = \delta_t \cdot \mu. \tag{10}$$

where $\mu \in (0,1)$.

To address the low optimization accuracy of a BAS algorithm, we incorporate the cubic interpolation into its updating rule. Firstly, we define a third-order polynomial, which is approximately equal to the objective function.

$$g(w) = c_0 + c_1 w + c_2 w^2 + c_3 w^3 = f(w), \tag{11}$$

where $c_0$, $c_1$, $c_2$, $c_3$ represent constants. The first derivative of the polynomial $g(w)$ is defined as zero, then we can achieve the following equation

$$g'(w) = c_1 + 2c_2 w + 3c_3 w^2 = 0, \tag{12}$$

Based on equation (12), we obtain that

$$w' = \frac{1}{3c_3}\left(-c_2 \pm \sqrt{c_2^2 - 3c_1 c_3}\right), \tag{13}$$

By combining equations (11) and (12), we obtain that

$$\begin{cases} g(w_1) = c_0 + c_1 w_1 + c_2 w_1^2 + c_3 w_1^3 = f(w_1) = f_1, \\ g(w_2) = c_0 + c_1 w_2 + c_2 w_2^2 + c_3 w_2^3 = f(w_2) = f_2, \\ g(w_3) = c_0 + c_1 w_3 + c_2 w_3^2 + c_3 w_3^3 = f(w_3) = f_3, \\ g'(w_1) = c_1 + 2c_2 w_1 + 3c_3 w_1^2 = f'(w_1) = f_1'. \end{cases} \tag{14}$$

From equations (11)-(14), we actually have the following equation.

$$\begin{cases} c_3 = \frac{\beta - \chi}{\kappa - \varphi}, \\ c_2 = \beta - \kappa c_3, \\ c_1 = f_1' - 2c_2 w_1 - 3c_3 w_1^2. \end{cases} \Rightarrow \begin{cases} \beta = \frac{f_2 - f_1 + f_1' \cdot (w_1 - w_2)}{(w_1 - w_2)^2}, \\ \chi = \frac{f_3 - f_1 + f_1' \cdot (w_1 - w_3)}{(w_1 - w_3)^2}, \\ \kappa = \frac{2w_1^2 - w_2(w_1 + w_2)}{(w_1 - w_2)}, \\ \varphi = \frac{2w_1^2 - w_3(w_1 + w_3)}{(w_1 - w_3)}. \end{cases} \tag{15}$$

where $w_1$, $w_2$, $w_3$ are three interpolation position. Based on the above analysis, we can achieve the minimum point $w'$.

*B. PF Algorithm*

Considering the influence of measurement noises and the highly nonlinear kinematic model, PF algorithm is adopted to address this critical issue [11-15]. Additionally, we discuss the PF algorithm in this section, whose state transition equation can be represented as

$$w_k = w_{k-1} + V_k, \tag{16}$$

$$Y_k = K(w_N + w_k) - K(w_N), \tag{17}$$

$$K(w_N) = T_6 = {}^0T_1\,{}^1T_2\,{}^2T_3\,{}^3T_4\,{}^4T_5\,{}^5T_6, \tag{18}$$

$$K(w_N + w_k) = T_6 + dT_6. \tag{19}$$

where $V_k$, $Y_k$, $K$ are the system noise, position error of robot arm, the forward kinematic operator, respectively.

Thereafter, we can obtain the particle updated equation as follows.

$$w_k^i = w_{k-1}^i + V_k, \tag{20}$$

By combining (17) and (20), we have

$$Y_k^i = K(w_N + w_k^i) - K(w_N) \tag{21}$$

Based on calibration rule, we achieve the weight of particles.

$$\rho_k^i = \frac{1}{\sqrt{2\pi|R|}} \exp\left(-\frac{1}{2}\left[Z_k - L_k^i\right]^T R^{-1}\left[Z_k - L_k^i\right]\right) \tag{22}$$

$Z_k$, $R$ and $L_k^i$ are the error of cable length measured by the drawstring displacement sensor, the covariance matrix of the measurement noises and the error of nominal cable length, respectively. From the equation (22), the normalization of the particle weight can be calculated by

$$\tilde{\rho}_k^i = \frac{\rho_k^i}{\sum_{j=1}^{N} \rho_k^i} \tag{23}$$

$$w_k = \sum_{i=1}^{N} \tilde{\rho}_k^i w_k^i \tag{24}$$

TABLE I
THE CALIBRATION ACCURACY OF M1-M10.

| Item | RMSE(mm) | Std(mm) | Max(mm) |
|---|---|---|---|
| Before | 2.09 | 2.00 | 3.36 |
| M1 | 0.66 | 0.56 | 1.71 |
| M2 | 0.70 | 0.56 | 1.76 |
| M3 | 0.64 | 0.55 | 1.51 |
| M4 | 0.59 | 0.49 | 1.28 |
| M5 | 0.85 | 0.72 | 1.91 |
| M6 | 0.50 | 0.41 | 1.16 |
| M7 | 0.67 | 0.58 | 1.59 |
| M8 | 0.82 | 0.67 | 1.79 |
| M9 | 0.51 | 0.43 | 1.27 |
| M10 | 0.48 | 0.39 | 1.08 |

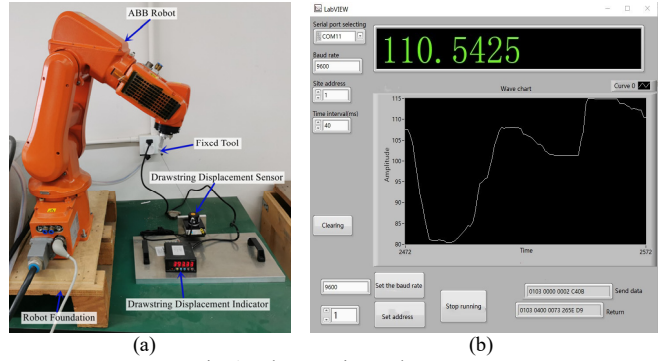

(a)     (b)

Fig. 1. The experimental system.

## IV. CALIBRATION EXPERIMENTS

### A. General Settings

*1) Evaluation Protocol:* In this work, we adopt the root mean square error (RMSE), average error (Std) and maximum error (Max) as the evaluation protocol [41-45].

$$Max = \max\left\{\sqrt{(L_i - L_i')^2}\right\}, Std = \frac{1}{n}\sum_{i=1}^{n}\sqrt{(L_i - L_i')^2}, \ RMSE = \sqrt{\frac{1}{n}\sum_{i=1}^{n}(L_i - L_i')^2}, \ i = 1, 2, \cdots n \tag{25}$$

*2) Dataset:* We randomly select 120 measurement points in the workspace of ABB IRB120 robot arm.

*3) Experimental Setup:* Fig. 1 shows the robot arm calibration experimental system, which includes an ABB IRB120 industrial robot arm, a drawstring displacement indicator, RS485 communication module, a drawstring displacement sensor and a PC.

### B. Comparison Experiments

*1) Compared With Advanced Methods*

Six state-of-the-art calibration methods are compared with the proposed method, whose detailed summaries are as follows:
a) **M1: Extended Kalman filter (EKF) algorithm** [28]. It is a direct method to address the issue of nonlinear system.
b) **M2: Beetle antennae search (BAS) algorithm**, which is developed in [29]. It is a biologically inspired intelligent optimization algorithm, which can achieve the efficient optimization without knowing the specific form of function and gradient information.
c) **M3: Unscented Kalman filter algorithm (UKF)** [30], which adopts traceless transformation to sample the state value.
d) **M4: Particle swarm optimization algorithm (PSO)**, which is presented in [31]. It is inspired by the foraging behavior of birds.
e) **M5: Radial basis function neural network (RBF)**, which is discussed in [32].
f) **M6: Levenberg-Marquardt (LM) algorithm** [3]. It modifies the updating rule of the traditional least square method.
g) **M7: Differential evolution algorithm (DE)**, which is proposed in [33].
h) **M8: Particle filter algorithm (PF)** [10]. It is a nonlinear filtering algorithm.
i) **M9: Cubic interpolated beetle antennae search algorithm (CIBAS)**.
j) **M10**: A **PF-CIBAS** method proposed in this work.

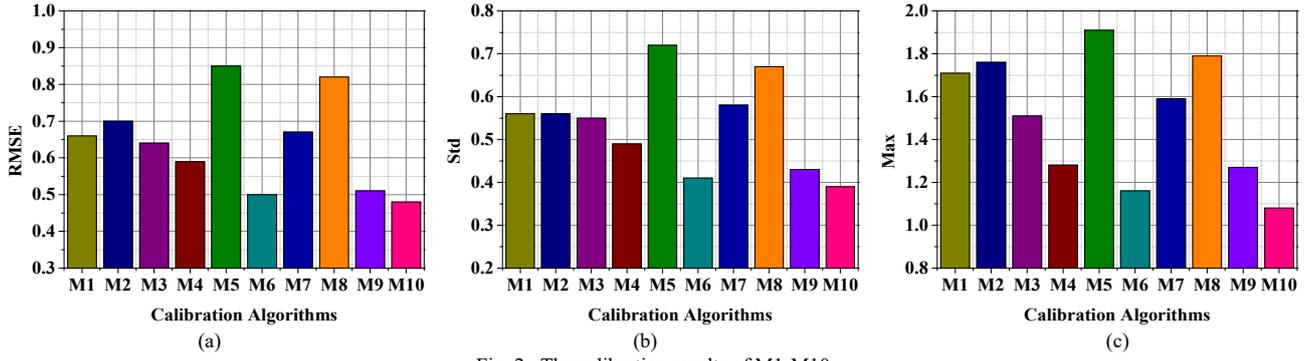

Fig. 2. The calibration results of M1-M10.

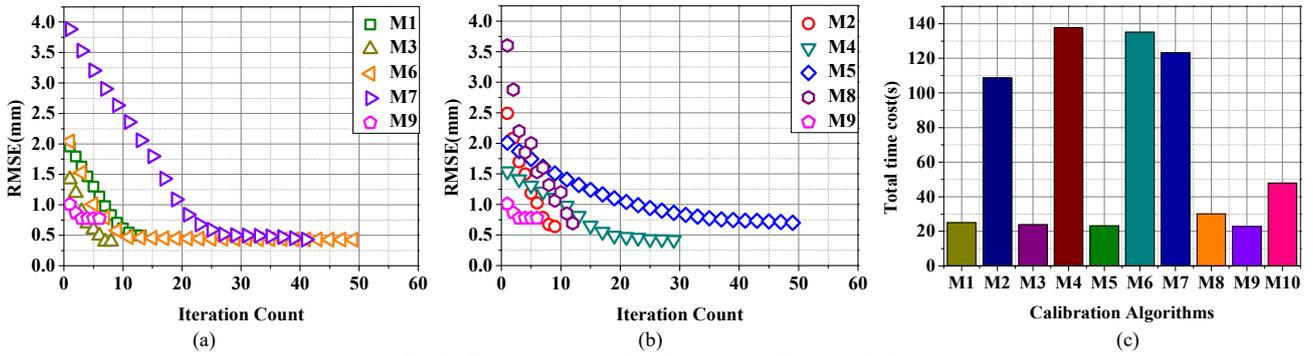

Fig. 3. Training curves and total time cost of these methods.

*2) Performance of Compared Methods*

In this part, we discuss the calibration performance of M1-10. After calibration, Table I and Fig. 2 show the calibration accuracy of M1-M10. Additionally, their training curves and total time costs are depicted in Fig. 3. From the above experiments, we find that

a) Compared with other calibration methods, M10 has the highest calibration accuracy. From Fig. 2, the RMSE, Std and Max of M10 are 0.48, 0.39 and 1.08, respectively. The most accurate method M6 achieves the RMSE, Std and Max of 0.50, 0.41 and 1.16 respectively, which is 4.17%, 5.13% and 7.41% higher than that of M10, respectively.

b) As depicted in Fig. 3(a) and (b), M9 obtain the fastest convergence speed, which only needs 2 iterations to converge in RMSE. However, the other fastest method M2 takes 7 iterations to converge in RMSE. From the above analysis, incorporating the cubic interpolation to BAS algorithm can obviously improve the convergence speed.

c) From Fig.3(c), M9 needs the least time cost, which only takes 22.83s to converge in RMSE. However, the other fastest M3 requires 23.85s, its time cost is 4.47% higher than that of M9.

## V. CONCLUSIONS

To improve the positioning accuracy of a robot arm, a new robot arm calibration method based on PF-CIBAS algorithm is proposed in this work. Compared with other methods, the proposed method achieves higher calibration accuracy.

REFERENCES

[1] M. S. Shang, Y. Yuan, X. Luo, and M. C. Zhou, "An *α-β*-divergence-generalized Recommender for Highly-accurate Predictions of Missing User Preferences," *IEEE Trans. on Cybernetics*, DOI: 10.1109/TCYB.2020.3026425.
[2] W. Wang, H. Song, Z. Yan, L. Sun, Z. Du, "A universal index and an improved PSO algorithm for optimal pose selection in kinematic calibration of a novel surgical robot," *Robot. Comput.-Integr. Manuf.*, vol. 50, pp. 90-101, Apr. 2018.
[3] Y. Gan, J. Duan, X. Dai, "A calibration method of robot kinematic parameters by drawstring displacement sensor," *Int. J. Adv. Robot. Syst*, vol.16, no.5, pp.1-9, Sep.-Oc. 2019.
[4] G. L. Wu and G. L. Shi, "Experimental statics calibration of a multiconstraint parallel continuum robot," *Mech. Mach. Theory*, vol. 136, pp. 72-85, Jun. 2019.
[5] X. Luo, Z. Ming, Z. H. You, S. Li, Y. N. Xia, and H. Leung, "Improving network topology-based protein interactome mapping via collaborative filtering," *Knowledge-Based Systems*, vol. 90, pp. 23-32, 2015.
[6] X. Wen, C. Kang, G. Qiao, D. Wang, Y. Han, A. Song, "Study on robot accuracy based on full pose measurement and optimization," *Chinese Journal of Scientific Instrument*, vol. 40, no. 7, pp. 81-89, Jul. 2019.
[7] X. Luo, Y. Yuan, M. C. Zhou, Z. G. Liu, and M. S. Shang, "Non-negative Latent Factor Model based on *β*-divergence for Recommender Systems," *IEEE Trans. on System, Man, and Cybernetics: Systems*, vol. 51, no. 8, pp. 4612-4623, 2021.
[8] H. Wu, X. Luo, and M. C. Zhou, "Advancing Non-negative Latent Factorization of Tensors with Diversified Regularizations," *IEEE Trans. on Services Computing*, DOI: 10.1109/TSC.2020.2988760, 2020.

[9] Y. Wang, J. Huang, DF. Wu, Z. Ma, "Calibration of an Industry Robot and External Axle Based on Ant Colony Optimization," *Advanced Materials Research*, vol. 301-303, pp. 1782-1788, Dec. 2011.
[10] X. Deng, L. Ge, R. Li and Z. Liu, "Research on the kinematic parameter calibration method of industrial robot based on LM and PF algorithm," *in Proc. of the 32th Int. Conf. on Chinese Control and Decision*, Hefei, China, Aug. 2020, pp. 2198-2203.
[11] X. Luo, Z. G. Liu, S. Li, M. S. Shang, and Z. D. Wang, "A Fast Non-negative Latent Factor Model based on Generalized Momentum Method," *IEEE Trans. on System, Man, and Cybernetics: Systems*, vol. 51, no. 1, pp. 610-620, 2021.
[12] Z. G. Liu, X. Luo, and Z. D. Wang, "Convergence Analysis of Single Latent Factor-Dependent, Nonnegative, and Multiplicative Update-Based Nonnegative Latent Factor Models," *IEEE Trans. on Neural Networks and Learning Systems*, vol. 32, no. 4, pp. 1737-1749, 2021.
[13] Y. Yuan, Q. He, X. Luo, and M. S. Shang, "A multilayered-and-randomized latent factor model for high-dimensional and sparse matrices," *IEEE Trans. on Big Data*, DOI: 10.1109/TBDATA.2020.2988778, 2020.
[14] Jiang. X, Li. S, "BAS: Beetle antennae search algorithm for optimization problems," *International Journal of Robotics & Control*, vol. 1, no.1, pp. 1-3, Jul. 2018.
[15] M. Yazdani, H. Salarieh, and M. S. Foumani, "Bio-inspired decentralized architecture for walking of a 5-link biped robot with compliant knee joints," *Int. J. Control Autom. Syst.*, vol. 16, no. 6, pp. 2935-2947, 2018.
[16] X. Luo, Y. Zhou, Z. G. Liu, L. Hu, and M. C. Zhou, "Generalized Nesterov's Acceleration-incorporated, Non-negative and Adaptive Latent Factor Analysis," *IEEE Trans. on Services Computing*, DOI: 10.1109/TSC.2021.3069108, 2021.
[17] J. Carff, M. Johnson, EM. El-Sheikh, "Human-robot team navigation in visually complex environments," *in Proc. of 2009 IEEE/RSJ Int. Conf. on Intelligent Robots and Systems*, St Louis, MO, USA, Oct. 2009, pp. 3043-3050.
[18] X. Luo, Z. D. Wang, and M. S. Shang, "An Instance-frequency-weighted Regularization Scheme for Non-negative Latent Factor Analysis on High Dimensional and Sparse Data," *IEEE Trans. on System Man Cybernetics: Systems*, vol. 51, no. 6, pp. 3522-3532, 2021.
[19] D. Wu, M. S. Shang, X. Luo, and Z. D. Wang, "An $L_1$-and-$L_2$-norm-oriented Latent Factor Model for Recommender Systems, *IEEE Trans. on Neural Networks and Learning Systems*, DOI: 10.1109/TNNLS.2021.3071392, 2021.
[20] H. Zhuang, Z.S. Roth, F. Hamano, "A complete and parametrically continuous kinematic model for robot manipulators," *IEEE Trans. Robot. Autom*, vol. 8, no.4, pp. 451-463, Aug. 1992.
[21] D. Wu, Y. He, X. Luo, and M. C. Zhou, "A Latent Factor Analysis-based Approach to Online Sparse Streaming Feature Selection," *IEEE Trans. on System Man Cybernetics: Systems*, DOI: 10.1109/TSMC.2021.3096065, 2021.
[22] A. Joubair and A. Bonev, "Comparison of the efficiency of five observability indices for robot calibration," *Mech. Mach. Theory*, vol. 70, no. 6, pp. 254-265, Dec. 2013.
[23] X. Luo, Z. G. Liu, M. S. Shang, J. G. Lou, and M. C. Zhou, "Highly-Accurate Community Detection via Pointwise Mutual Information-Incorporated Symmetric Non-negative Matrix Factorization," *IEEE Trans. on Network Science and Engineering*, vol. 8, no. 1, pp. 463-476, 2021.
[24] L. Hu, P. W. Hu, X. Y. Yuan, X. Luo, and Z. H. You, "Incorporating the Coevolving Information of Substrates in Predicting HIV-1 Protease Cleavage Sites," *IEEE/ACM Trans. on Computational Biology and Bioinformatics*, vol. 17, no. 6, pp. 2017-2028, 2020.
[25] Z. Li, S. Li, O. O. Bamasag, A. Alhothali and X. Luo, "Diversified Regularization Enhanced Training for Effective Manipulator Calibration," *IEEE Trans. Neural Networks and Learning Systems*, DOI: 10.1109/TNNLS.2022.3153039.
[26] X. Luo, W. Qin, A. Dong, K. Sedraoui, and M. C. Zhou, "Efficient and High-quality Recommendations via Momentum-incorporated Parallel Stochastic Gradient Descent-based Learning," *IEEE/CAA Journal of Automatica Sinica*, vol. 8, no. 2, pp. 402-411, 2021.
[27] X. Luo, Y. Yuan, S. L. Chen, N. Y. Zeng, and Z. D. Wang, "Position-Transitional Particle Swarm Optimization-Incorporated Latent Factor Analysis," IEEE Trans. on Knowledge and Data Engineering, DOI: 10.1109/TKDE.2020.3033324, 2020
[28] Z. H. Jiang, W. G. Zhou, H. Li, Y. Mo, W. C. Ni, and Q. Huang, "A new kind of accurate calibration method for robotic kinematic parameters based on the extended Kalman and particle filter algorithm," *IEEE Trans. Ind. Electron.*, vol. 65, no. 4, pp. 3337-3345, Apr. 2018.
[29] Y. X. Wang, Z. W. Chen, H. F. Zu, X. Zhang, C. T. Mao, and Z. R. Wang, "Improvement of heavy load robot positioning accuracy by combining a model-based identification for geometric parameters and an optimized neural network for the compensation of nongeometric errors," *Complexity*, vol. 2020, Article ID 5896813, Jan. 2020.
[30] G. Du, Y. Liang, C. Li, P. X. Liu and D. Li, "Online robot kinematic calibration using hybrid filter with multiple sensors," *IEEE Trans. on Instrumentation and Measurement*, vol. 69, no. 9, pp. 7092-7107, Sept. 2020.
[31] J. H. Lee, J. Song, D. Kim, J. Kim, Y. Kim and S. Jung, "Particle swarm optimization algorithm with intelligent particle number control for optimal design of electric machines," *IEEE Trans. Ind. Electron.*, vol. 65, no. 2, pp. 1791-1798, Feb. 2018.
[32] D. Chen, T. M. Wang, P. J. Yuan, et al., "A positional error compensation method for industrial robots combining error similarity and radial basis function neural network," *Meas. Sci. Technol.*, vol. 30, no. 12, pp. 125010, Sep. 2019.
[33] S. Zhou, L. Xing, X. Zheng, N. Du, L. Wang and Q. Zhang, "A self-adaptive differential evolution algorithm for scheduling a single batch-processing machine with arbitrary job sizes and release times," *IEEE Trans. on Cybernetics*, vol. 51, no. 3, pp. 1430-1442, Mar. 2021.
[34] A. Joubair and I. A. Bonev, "Non-kinematic calibration of a six-axis serial robot using planar constraints," *Precis. Eng.*, vol. 40, pp.325-333, Apr. 2015.
[35] T. S. Lembono, F. Suárez-Ruiz, and Q. C. Pham, "SCALAR: Simultaneous calibration of 2-D laser and robot kinematic parameters using planarity and distance constraints," *IEEE Trans. Autom. Sci. Eng.*, vol. 16, no. 4, pp. 1971-1979, Oct. 2019.
[36] X. Luo, M. Zhou, S. Li, D. Wu, Z. Liu and M. Shang, "Algorithms of Unconstrained Non-Negative Latent Factor Analysis for Recommender Systems," *IEEE Trans. on Big Data*, vol. 7, no. 1, pp. 227-240, 1 Mar. 2021.
[37] D. Wu, X. Luo, M. Shang, Y. He, G. Wang and M. Zhou, "A deep latent factor model for high-dimensional and sparse matrices in recommender systems," *IEEE Trans. on System, Man, and Cybernetics: Systems*, vol. 51, no. 7, pp. 4285-4296, Jul. 2021.
[38] X. Luo, M. Zhou, Z.-D. Wang, Y. Xia, Q. Zhu, "An effective scheme for QoS estimation via alternating direction method-based matrix factorization," *IEEE Trans. on Services Computing*, vol. 12, no. 4, pp. 503-518, Jul. 2019.
[39] X. Luo, D. Wang, M. Zhou and H. Yuan, "Latent factor-based recommenders relying on extended stochastic gradient descent algorithms," *IEEE Trans. on System, Man, and Cybernetics: Systems*, vol. 51, no. 2, pp. 916-926, Feb. 2021.
[40] Y. Jiang, T. M. Li, L. P. Wang, and F. F. Chen, "Kinematic error modeling and identification of the over-constrained parallel kinematic machine," *Robot. Comput.-Integr. Manuf*, vol. 49, pp. 105-119, Feb. 2018.
[41] X. Luo, H. Wu, Z. Wang, J. Wang and D. Meng, "A novel approach to large-Scale dynamically weighted directed network representation," *IEEE Trans. on Pattern Analysis and Machine Intelligence*, DOI: 10.1109/TPAMI.2021.3132503.
[42] S. L. Zhang, S. Wang, F. S. Jing, and M. Tan, "A sensorless hand guiding scheme based on model identification and control for industrial robot," *IEEE Trans. Ind. Inform.*, vol. 15, no. 9, pp. 5204-5213, Sep. 2019.
[43] C. T. Mao, S. Li, Z. W. Chen, X. Zhang, and C. Li, "Robust kinematic calibration for improving collaboration accuracy of dual-arm manipulators with experimental validation," *Measurement*, vol. 155, pp. 107524, Apr. 2020.
[44] Z. Li, S. Li, and X. Luo, "An overview of calibration technology of industrial robots," *IEEE/CAA J. Autom. Sinica*, vol. 8, no. 1, pp. 23-36, Jan. 2021.
[45] S. Li, J. He, Y. Li and M. U. Rafique, "Distributed recurrent neural networks for cooperative control of manipulators: A game-theoretic perspective," *IEEE Trans. Neural Netw. Learn. Syst.*, vol. 28, no. 2, pp. 415-426, Feb. 2017.